\documentclass{article}

\PassOptionsToPackage{numbers, compress}{natbib}

\usepackage[final]{neurips_wrl2023}

\usepackage[utf8]{inputenc} 
\usepackage[T1]{fontenc}    
\usepackage{hyperref}       
\usepackage{url}            
\usepackage{booktabs}       
\usepackage{amsfonts}       
\usepackage{nicefrac}       
\usepackage{microtype}      
\usepackage{graphicx}

\usepackage{subcaption}
\usepackage{amsmath}

\title{Reinforcement-learning robotic sailboats: simulator and preliminary results}

\author{%
  Eduardo C.~Vasconcellos
    \\
  Fluminense Federal University\\
  Niter\'oi, RJ, Brazil \\
  \texttt{evasconcellos@id.uff.br} \\
  \And
  Ronald M.~Sampaio \\
  Fluminense Federal University\\
  Niter\'oi, RJ, Brazil \\
  \texttt{ronaldmaymone@id.uff.br} \\
  \AND
  Andr\'e P.D. Ara\'ujo \\
  Fluminense Federal University\\
  Niter\'oi, RJ, Brazil \\
  \texttt{andrepda@id.uff.br} \\
  \And
  Esteban W.~G.~Clua \\
  Fluminense Federal University\\
  Niter\'oi, RJ, Brazil \\
  \texttt{esteban@ic.uff.br}\\
  \And
  Philippe Preux \\
  Center Inria de l'Universit\'e de Lille \\
  Villeneuve d'Ascq, Lille, France \\
  \texttt{philippe.preux@univ-lille.fr} \\
  \And
  Raphael Guerra \\
  Fluminense Federal University\\
  Niter\'oi, RJ, Brazil \\
  \texttt{rguerra@ic.uff.br}
  \And
  Luiz M.~G.~Gon\c{c}alves \\
  Federal University of Rio Grande do Norte \\
  Natal, RN, Brazil\\
  \texttt{lmarcos@dca.ufrn.br}
  \And
  Luis Mart\'{i} \\
  Center Inria Chile \\
  Las Condes, Chile \\
  \texttt{lmarti@inria.cl}\\
  \And
  Hernan Lira \\
  Center Inria Chile \\
  Las Condes, Chile \\
  \texttt{hernan.lira@inria.cl}\\
  \And
  Nayat S.~Pi \\
  Center Inria Chile \\
  Las Condes, Chile \\
  \texttt{nayat.sanchez-pi@inria.cl}\\
}

\begin{document}

\maketitle

\begin{abstract}

  This work focuses on the main challenges and problems in developing a virtual oceanic environment reproducing real experiments using Unmanned Surface Vehicles (USV) digital twins. We introduce the key features for building virtual worlds, considering using Reinforcement Learning (RL) agents for autonomous navigation and control. With this in mind, the main problems concern the definition of the simulation equations (physics and mathematics), their effective implementation, and how to include strategies for simulated control and perception (sensors) to be used with RL. We present the modeling, implementation steps, and challenges required to create a functional digital twin based on a real robotic sailing vessel. The application is immediate for developing navigation algorithms based on RL to be applied on real boats.
\end{abstract}

\section{Introduction}

The development of Unmanned Surface Vessels (USVs), also known as Autonomous Surface Vessels (ASVs), is currently an interesting and exciting research field, with applications in surveillance, rescue missions, structural inspection, and environmental monitoring, among others \cite{jorge:2019, silvajunior:2016, junior:2013}. These ASVs can be applied to fulfill dangerous or long recursive missions that could risk human lives or have a higher cost when executed by a human crew.

Developing a fully autonomous USV is not easy, and a significant challenge is the Guidance, Navigation, and Control (GNC) system \cite{liu:2016,paravisi:2019}. System prototyping and testing could be costly with a real USV. An undesired or unexpected behavior can damage the USV or other environmental entities. Therefore, it is usual that researchers fall back on simulated environments, where one can run different experiments without being afraid of losing the USV or causing damage to others. Simulation also has the advantage that one can control various aspects of it, performing several simultaneous experiments. For example, one can submit the USV to rare (or extreme) environmental conditions, which might not always be possible in the real environment.

A USV motion can be sourced by motorized propellers (most common boats), using the wind (sailboats), or even using a mix of them, which is the one that will be treated here. While other unmanned vehicles, like cars and many different types of robots, are often simulated using the physics of rigid bodies and collisions, a USV is under the influence of other forces generated by the water and air. For a USV simulation, one needs to compute not only the dynamic and static forces of a rigid body but also the hydrodynamic, hydrostatic, and aerodynamic forces acting on the hull. In the case of a sailing robot, aerodynamic forces from the wind also act on the sail's surface.

In this direction, this paper aims to present some challenges, problems, and a possible approach for building a virtual oceanic environment application, considering that it will be populated with USV digital twins, which include sailing robots. Our proposal is based on a literature review and on experiences in building real autonomous vessels and their simulated digital twin scenarios. The virtual world has been created based on our experience developing real USVs for environmental monitoring (including autonomous motor-propelled and sailboats). Its application is immediate, with the main goal of saving time and effort to develop navigation algorithms to be applied on our physical boats. Also, we are developing reinforcement learning algorithms and other machine learning tasks to be applied to our real USVs on top of this platform. Hence, the contribution of this work resides in introducing several aspects, challenges, and requirements necessary for creating the proposed digital twin. 

\section{Key features on building a reliable simulator for sailing robots}
\label{sect:APIs}

A simulated world for robotic development must implement physics, sensors, and actuators so the robot can perceive and interact with its environment. To enable the implementation of sensors like cameras and LiDAR, the simulator must be capable of rendering a realistic 3D scene representing the robot's surrounding environment. Therefore, a system with reliable physics and rendering capabilities is very desirable. Some authors advocate that modern game engines will be good candidates for creating realistic simulations, joining all characteristics that a developer desires \cite{juliani:2018}.

\subsection{Physics}\label{subsect:physcis_engines}

A physics engine is an important variable for building a reliable virtual world. Several physics engines can compute collision, contact, and reaction forces among rigid bodies.

On the other hand, water and wind simulations can be achieved using parametric models to describe the hydrodynamic, hydrostatic, and aerodynamic forces acting on different vessel parts (hull, keel, rudder, sail, etc). USV simulations were proposed using Gazebo ~\cite{paravisi:2019, bingham:2019} and Unity \cite{gan:2022}. Although, the only sailing robot simulator is within the work of Paravisi et al. (2019) \cite{paravisi:2019}.

 \subsection{ROS support}

 ROS (Robot Operational System) \cite{quigley:2009} is a largely used API within the robotics development community. ROS is a free, open-source set of libraries and tools that enables a developer or a team to integrate all systems that compose a robot, like motors, sensors, batteries, and control software. Implementing sensors and actuators is very important so the robot can perceive its environment and navigate through it. ROS enables us to build a communication infrastructure for collecting sensor data and sending commands to the actuators. Integrating the simulated environment with ROS makes commuting between a real and a virtual system easy.

 \subsection{Sensors and actuators}
\label{sect:sensors}

In a USV, the sensor array is essential to autonomous navigation and control. For so, they must be simulated as accurately as possible in a virtual world~\cite{Harris2011}. Depending on the simulation platform, one can find many implementations for various sensors, such as cameras, LiDAR, and GPS~\cite{Aranha:2002, Schluse:2018, paravisi:2019}. These implementations seek to reproduce actual sensors in all aspects, including noise, data structure, and transmission.

For a digital twin, the simulated sensor data is aimed to feed the navigation and control agent with information that describes the current state of the vessel. Then, the agent makes decisions, and through the actuators, it seeks to change the vessel's state. For a sailing robot, there are three basic actuators that the agent can control: the boom (which controls the main sail), the rudder (which controls the boat direction), and the propeller (forward propulsion, if available).

\subsection{Integration of USV digital twin into the environment}
\label{sect:omniverse}

This challenge is related to building the sailing robot's digital twin to emulate all the real boat's characteristics as accurately as possible, starting with creating a 3D model to represent the sailing robot in the virtual world.

\begin{figure}\centering
    \begin{subfigure}[t]{0.2\textwidth}\centering
        \includegraphics[width=\textwidth]{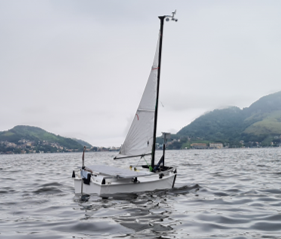}
        \caption{Real boat sailing.}
        \label{fig:realboat}
    \end{subfigure}\hfill
    \begin{subfigure}[t]{0.24\textwidth}\centering
        \centering
        \includegraphics[width=\textwidth]{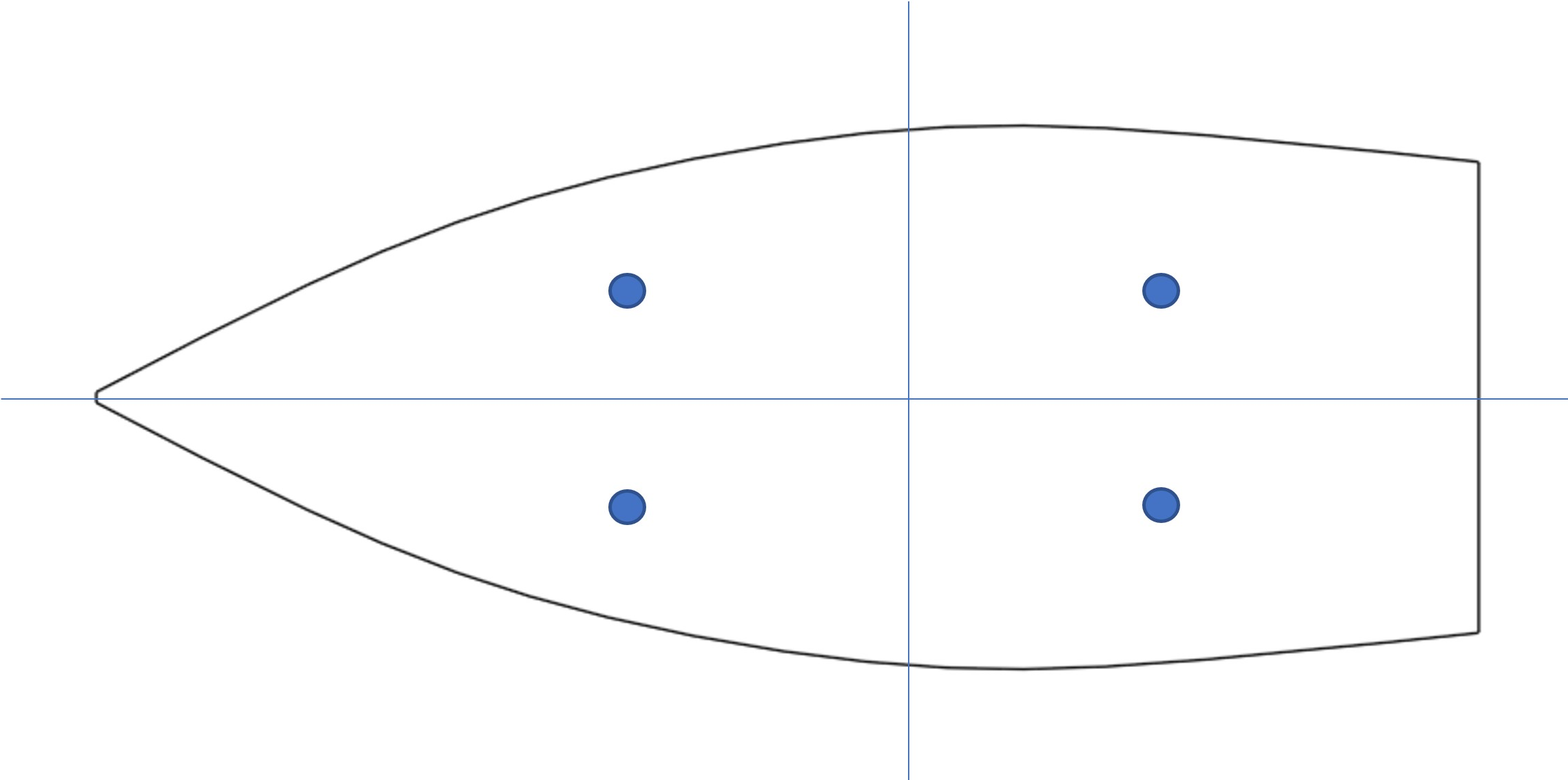}
        \caption{Hull division to apply the buoyancy force.}
        \label{fig:bottom_view_division}
    \end{subfigure}\hfill
    \begin{subfigure}[t]{0.13\textwidth}
        \centering
        \includegraphics[width=\textwidth]{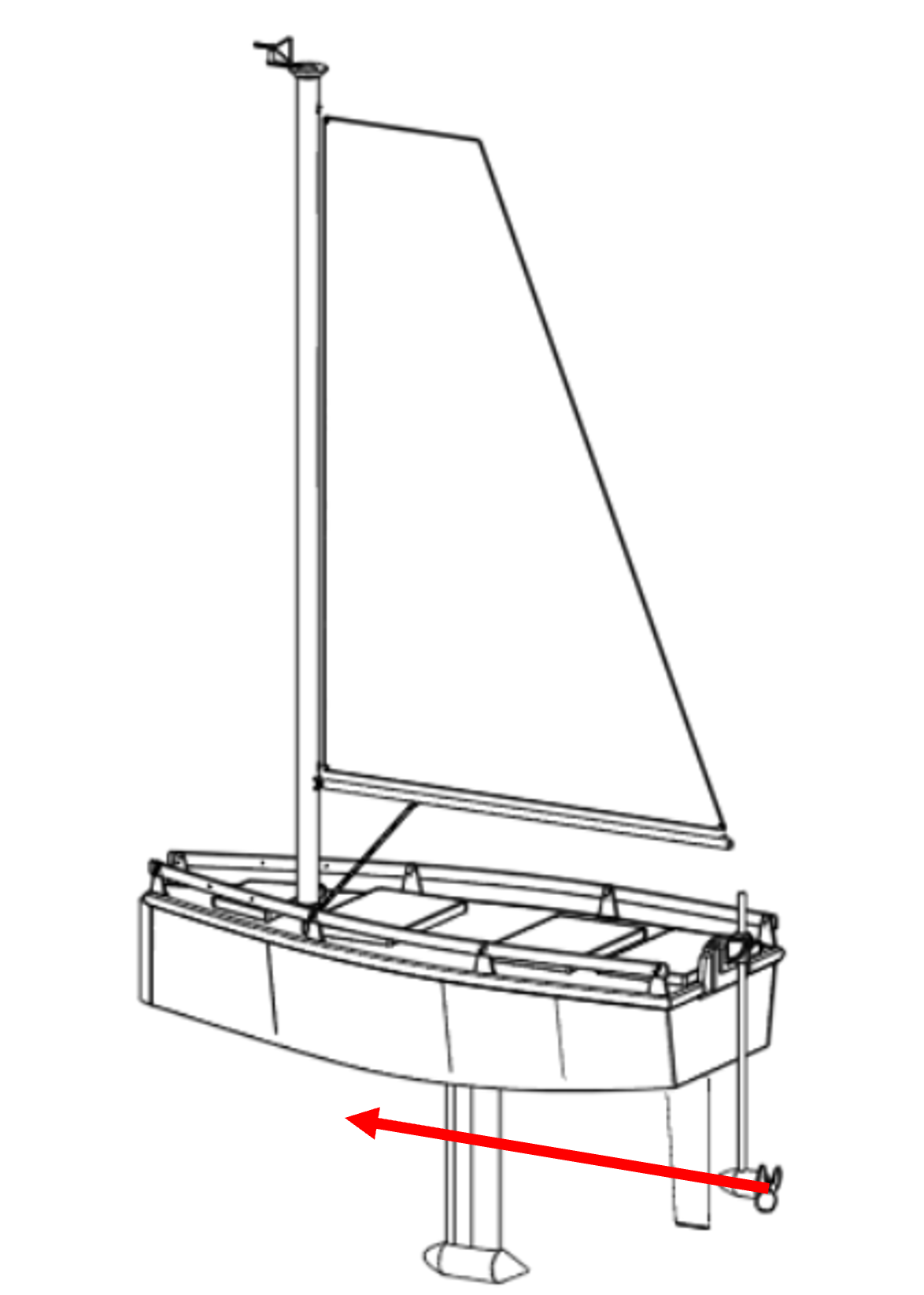}
        \caption{Propeller pushing force.}
        \label{fig:lift_propeller}
    \end{subfigure}\hfill
    \begin{subfigure}[t]{0.38\textwidth}\centering
        \includegraphics[width=\textwidth]{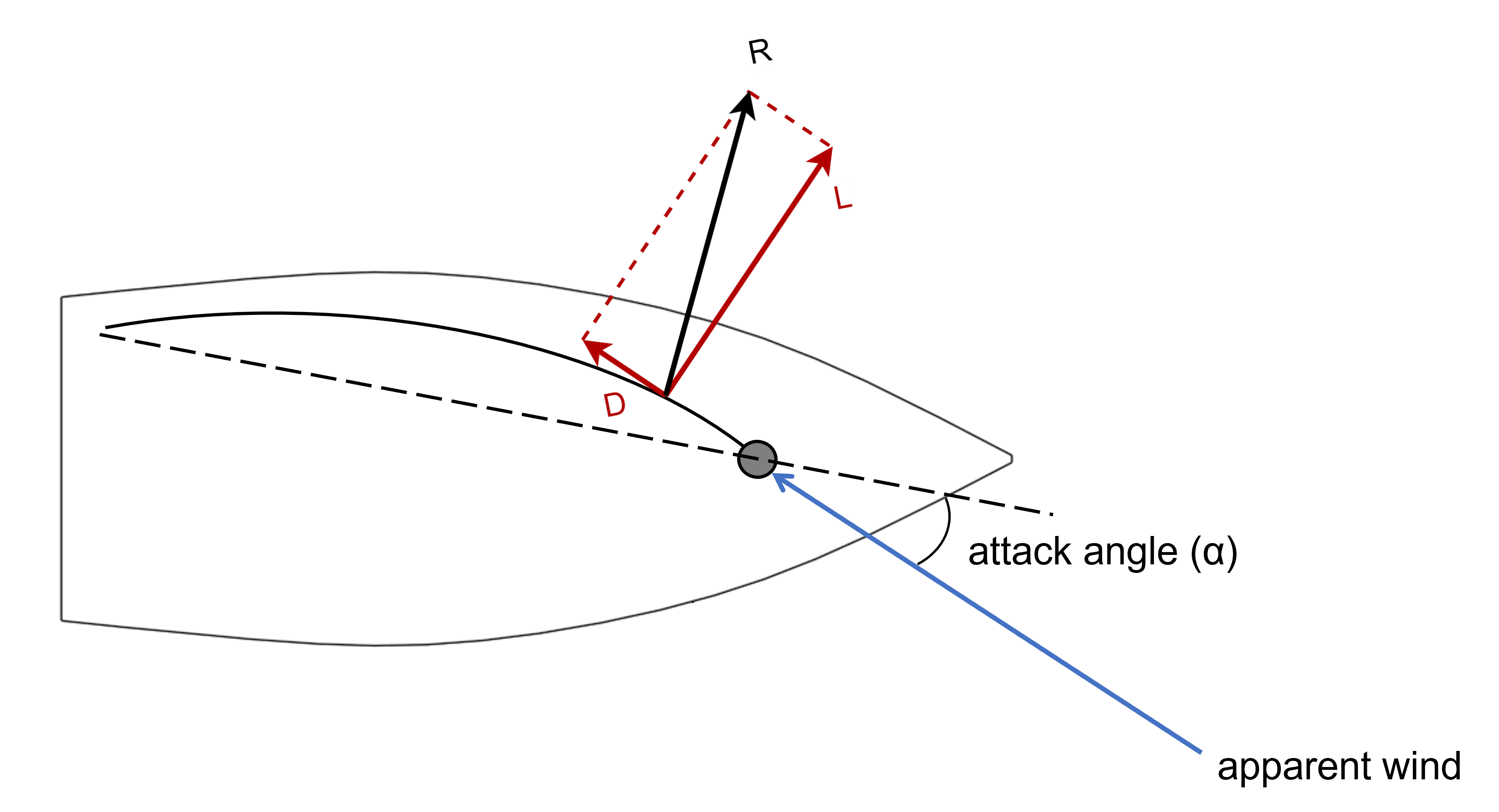}
        \caption{Aerodynamic lift and drag forces as a function of attack angle.}
        \label{fig:wind}
    \end{subfigure}
    \caption{E-Boat physics characterization.}
\end{figure}

We built a sailing robot digital twin, the E-Boat, to address this challenge. The 3D model, mass distribution, inertia matrix, and collision shapes were created using Onshape (https://www.onshape.com/), a Computer-aided Design (CAD) software delivered on the Internet using a software-as-a-service model. Nevertheless, this physics is just part of the whole environment~\cite{Anderson2003}. The boat motion under the influence of the water is modeled using the 6-DOF motion model proposed by Fossen~\cite{fossen:2011}, following an implementation strategy similar to the works of Paravisi~\cite{paravisi:2019} and Bingham~\cite{bingham:2019}. The hydrodynamic and hydrostatic forces are computed separately, and instead of computing a single buoyance force, we split the hull into four parts, calculating the buoyance force for each part as seen in Figure \ref{fig:bottom_view_division}. This solution helps to create a more realistic motion under wave conditions.

The sailing robot's main propulsion is the wind force, and we compute this force using equations:
\begin{equation}
    L = 0.5~\rho_\text{air} V_{A}^{2} ~A~C_{L}(\alpha);\quad D = 0.5~\rho_\text{air} V_{A}^{2}A~C_{D}(\alpha)\,. \label{eq:lift-drag}
\end{equation}
\noindent where $\rho_\text{air}$ is the air density, $V_{A}$ is the apparent wind speed, $A$ is the area of the sail, $\alpha$ is the attack angle, i.e., the angle between the sail chord line and the apparent wind direction, $C_{L}$ is the lift coefficient and $C_{D}$ is the drag coefficient. The $C_L(\alpha)$ and $C_D(\alpha)$ were estimated from real data from experimental missions performed with the sailing robot in the real world (see Figure~\ref{fig:realboat}). The propulsion force generated by the propeller is set by a discrete function described by the manufacturer and applied in the center of the propeller. Figure~\ref{fig:lift_propeller} illustrates the position where the force generated by the propeller is applied, and Figure~\ref{fig:wind} depicts the scheme for the wind forces acting on the center of effort of the sail.


We model the hydrodynamic forces acting on the rudder and the keel similarly we did for the sail. As the sailing robot moves, the water generates lift and drag forces on the keel and the rudder. The keel is static, but the rudder can change its attack angle and generate lift that will cause the robot to change direction. The equations used to calculate the forces are similar to equations in~(\ref{eq:lift-drag})
, where we replace the air density with the water density. The lift and drag coefficients $C_L(\alpha)$ and $C_D(\alpha)$ are also estimated by experiments with the real sailing robot.

\section{Towards autonomous navigation and control with reinforcement learning}

\begin{figure}[t]
    \centering\hfill
    \begin{minipage}[b]{0.35\textwidth}\centering
        \begin{subfigure}{\textwidth}\centering
            \includegraphics[width=\textwidth]{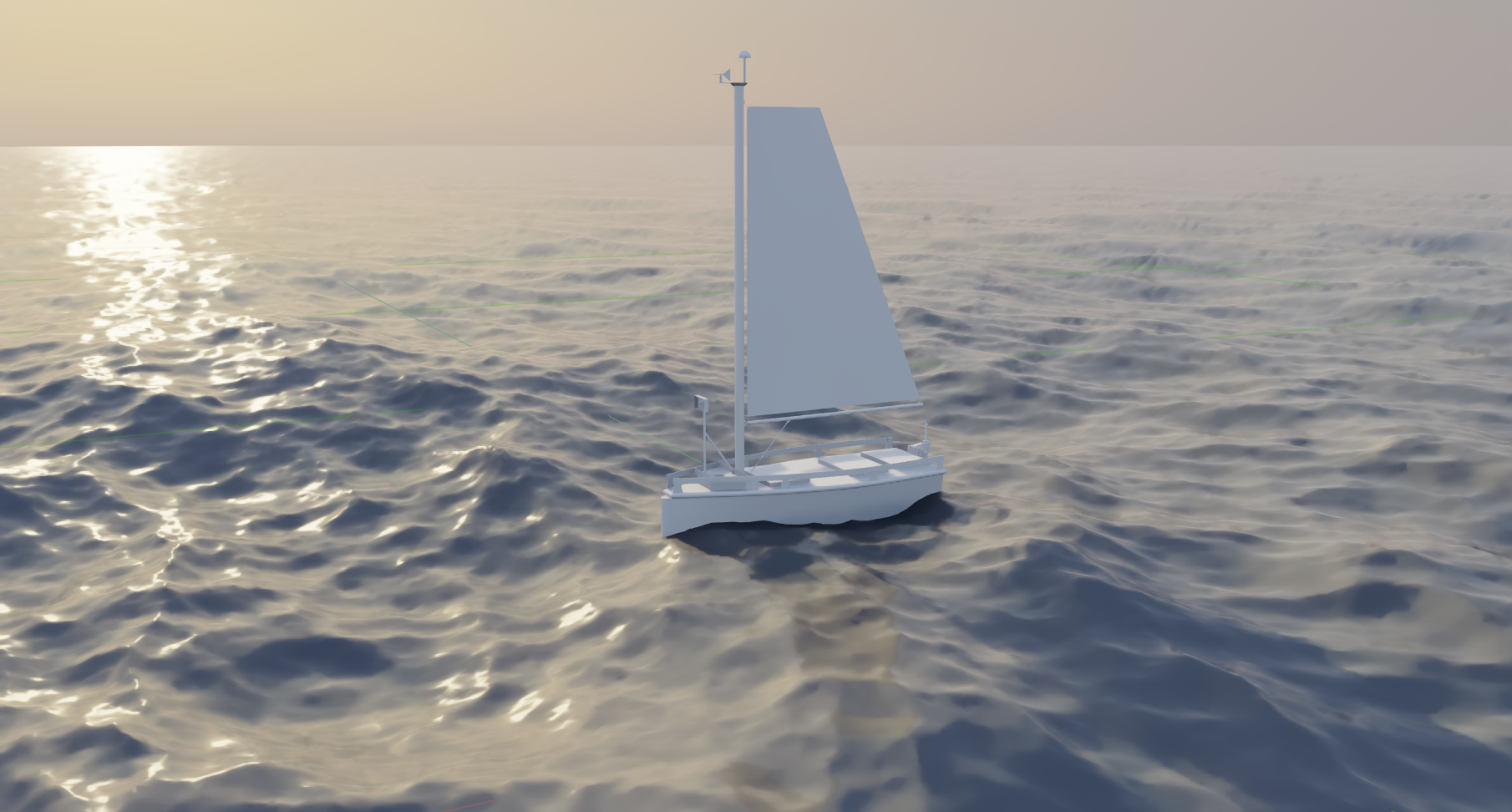}
            \caption{Omniverse rendering.}
        \end{subfigure}\\
        \begin{subfigure}{\textwidth}\centering
            \includegraphics[width=\textwidth]{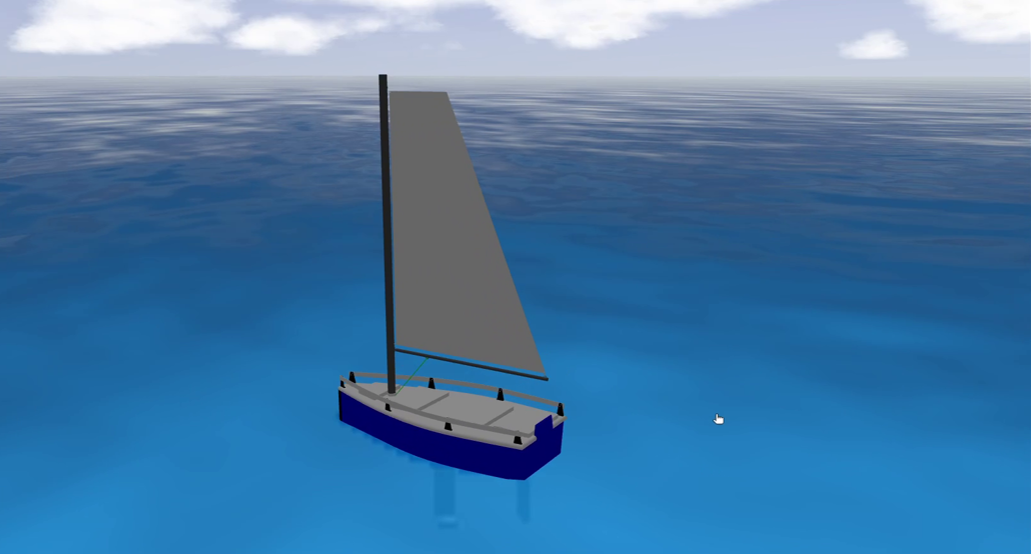}
            \caption{Gazebo rendering.}
        \end{subfigure}
        \caption{Sea surface rendering in Omniverse and Gazebo.}
        \label{fig:sea_surface}
    \end{minipage}\hfill
    \begin{minipage}[b]{0.55\textwidth}
        \centering
        \includegraphics[width=\textwidth]{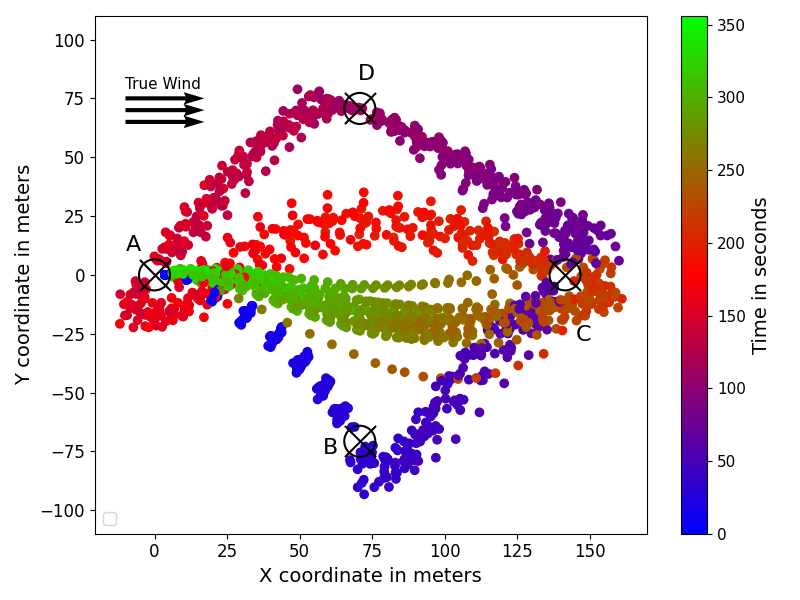}
        \caption{RL agent policy behavior in twenty runs of the same mission. The mission path is defined as BCDACA.}
        \label{fig:results}
    \end{minipage}\hfill
\end{figure}

Using the E-Boat and an open ocean environment we built based on the project VRX \cite{bingham:2019}, we conducted some experiments in autonomous GNC using reinforcement learning. Our main simulation platform is Gazebo~\cite{koenig:2004}. It offers a reliable physics engine, 
and the modularity to add sensors, actuators, and new physics to the simulation using C++ plugins. We are also developing an open-world simulation for Nvidia Omniverse. Omniverse has more robust physics and graphics engines, but as a broad tool to create virtual worlds, it is more complex and demands more computational power. Figure \ref{fig:sea_surface} depicts the sea surface rendering in Omnivers and Gazebo 11.

Using the Gazebo simulation combined with stable-baselines3 and Gymnasium, we managed to train an agent to navigate the E-Boat through a mission composed of six waypoints. The agent was trained using PPO \cite{schulman:2017}.

Figure \ref{fig:results} shows the progress of the E-Boat in twenty runs of the same mission. The mission path is defined as BCDACA. This initial result indicates the maneuvers executed by the agent are consistent, with small variations through the runs. The variation in the path from one run to another is due to the waves hitting the boat during the mission.

\section{Conclusions}
\label{conclusions}

There are many challenges to developing autonomous GNC systems for sailing robots. We believe that pursuing a reliable simulated environment is a key resource to address this challenge.

We have been developing an oceanic virtual environment mainly to support research in sailing robots, which will also support other types of USVs. We are focusing our efforts on Gazebo and Omniverse, due to their robust physics engines and connections with ROS and reinforcement learning APIs, like stable-baselines3. Although Omniverse is superior in rendering capabilities and offers a connection with a broad specter of software and development tools, Gazebo offers a more direct approach to building worlds for robotic simulations as it was conceived as a tool for the robotic community.

Our initial results with an RL agent to control and navigate the sailing robot were promising. In the future, we will develop this agent and compare its results with other algorithms in the literature.

\subsubsection*{Acknowledgments}

We acknowledge and are thankful to CAPES AMSUD (proj. 88881.522966/2020-01), NVIDIA, CNPq-FNDCT-MCTI (405535/2022-8) and the City of Niterói (Brazil) for supporting and funding this research.

\bibliographystyle{apalike}
\bibliography{manuscript.bib}

\end{document}